\documentclass[10pt,twocolumn,letterpaper]{article}

\usepackage{cvpr}              %

\definecolor{cvprblue}{rgb}{0.21,0.49,0.74}
\usepackage[pagebackref,breaklinks,colorlinks,allcolors=cvprblue]{hyperref}
\usepackage{multirow}
\usepackage{makecell}
\usepackage{xcolor} %
\def\paperID{15} %
\def\confName{CVPR}
\def\confYear{2026}

\title{InterventionLens: A Multi-Agent Framework for Detecting ASD Intervention Strategies in Parent-Child Shared Reading}

\author{%
Xiao Wang$^{*}$, Lu Dong$^{*}$, Ifeoma Nwogu, Srirangaraj Setlur, Venu Govindaraju\\
State University of New York at Buffalo, Buffalo, NY 14260, USA\\
{\tt\small \{xwang277, ludong, inwogu, setlur, govind\}@buffalo.edu}
\thanks{These authors contributed equally.}
}
\begin{document}
\maketitle
\begin{abstract}
Home-based interventions like parent-child shared reading provide a cost-effective approach for supporting children with autism spectrum disorder (ASD). 
However, analyzing caregiver intervention strategies in naturalistic home interactions typically relies on expert annotation, which is costly, time-intensive, and difficult to scale. 
To address this challenge, we propose InterventionLens, an end-to-end multi-agent system for automatically detecting and temporally segmenting caregiver intervention strategies from shared reading videos. 
Without task-specific model training or fine-tuning, InterventionLens uses a collaborative multi-agent architecture to integrate multimodal interaction content and perform fine-grained strategy analysis. Experiments on the ASD-HI dataset show that InterventionLens achieves an overall F1 score of 79.44\%, outperforming the baseline by 19.72\%. 
These results suggest that InterventionLens is a promising system for analyzing caregiver intervention strategies in home-based ASD shared reading settings. 
Additional resources will be released on the project page.
\end{abstract}
    
\section{Introduction}
\label{sec:intro}
Autism Spectrum Disorder (ASD) is a prevalent neurodevelopmental disorder that substantially impairs social communication and daily functioning \cite{american2013diagnostic,fuller2020effects}. While early intervention can improve core symptoms \citep{estes2015long}, many families lack consistent access to effective services due to long-term costs and uneven resource distribution \citep{lin2025inequality}. This gap has motivated scalable, caregiver-deliverable home-based interventions \cite{lord2022lancet,world2022training,wong2022adapting,wetherby2018changing}, among which shared book reading (SBR) is a particularly effective and accessible routine \citep{boyle2019effects,westerveld2021investigating}. In practice, families often record SBR sessions at home and submit the videos to speech-language pathologists (SLPs), who manually conduct video-by-video review to identify caregiver intervention strategies and provide guidance \cite{akemoglu2022parent}. However, such expert-driven annotation is labor-intensive, costly, and difficult to scale. Therefore, a key technical need is to automatically detect intervention strategies from home-based videos. In this work, we focus on three caregiver intervention strategies used in parent-implemented ASD interventions: \textit{Modeling}, where the caregiver demonstrates the target word or phrase for the child; \textit{Mand-Model}, where the caregiver prompts the child to respond and then provides a model when needed; and \textit{Time Delay}, where the caregiver deliberately pauses to create an opportunity for the child to initiate or complete the response \cite{meadan2013coaching,meadan2014parent,meadan2016internet}. These strategies originate from the Parent-Implemented Communication Strategies (PiCS) framework and were later adapted to shared reading and telepractice-based interventions for children with ASD \cite{akemoglu2022parent,akemouglu2022module,akemoglu2021parent}.

Achieving this goal is inherently difficult because shared book reading (SBR) is highly dynamic and interactive, typically involving a tightly coupled, triadic interaction among a caregiver, a child, and a book. Caregivers drive this process through continuous verbal scaffolding while simultaneously coordinating multi-modal social cues ---such as gaze, facial expressions, and gestures ---to regulate child engagement \cite{li2025asd}. Fine-grained recognition and reasoning of these interaction segments become even more challenging in unstructured, in-the-wild home settings. In these environments, non-fixed camera viewpoints, ambient noise, and variable interaction rhythms are compounded by the inherent acoustic differences between adult and child voices \citep{jain2023adaptationwhispermodelschild}. These factors drastically degrade the performance of mainstream Automatic Speech Recognition (ASR) systems like Whisper \cite{attia2024kid}, severely undermining the reliability of subsequent interaction modeling. Finally, these technical hurdles are further magnified by a severe scarcity of high-quality annotated data. Although the ASD-HI dataset \cite{li2025asd} partially addresses the gap in home-based SBR, its limited scale cannot support the massive data demands of modern data-intensive model training. 
Consequently, developing a robust and generalizable automated analysis system under these compounding constraints remains a core open challenge in the field.

To address these limitations, we propose InterventionLens, a hierarchical multi-agent framework for caregiver intervention detection without parameter updating. Our contributions are three-fold:

\begin{itemize}
    \item 
    We propose a two-layer multi-agent architecture that decouples multimodal perception from intervention detection and segmentation, combining structured transcript and book-context modeling with coarse-to-fine expert-based temporal localization.

    \item 
    We introduce a progressive knowledge refinement mechanism that uses a small amount of labeled data to iteratively update decision rules and refine the agent’s structured guidance. By translating clinical standard operating procedures (SOPs) into explicit agent roles and decision constraints, our framework enables structured strategy analysis without data-intensive fine-tuning.

    \item 
    We demonstrate strong empirical performance on the ASD-HI dataset, where InterventionLens substantially outperforms existing baselines under strict evaluation criteria, improving the overall F1 score from 59.72\% to 79.44\% and Precision from 50.21\% to 82.36\%.
\end{itemize}

\section{Related Work}
\label{sec:related_work}

\paragraph*{LLM-based Multi-Agent Systems:}
Large language model (LLM)-based multi-agent systems (MAS) have emerged as an effective paradigm for complex reasoning by decomposing problems into modular subtasks handled by specialized agents \cite{guo2024large,yang2024llm,tran2025multi,dong2024ig3d,dong2024word,dong2024signavatar}. Compared with single-agent systems, which often suffer from context dilution and error accumulation over long interactions, MAS can better isolate subproblems, reduce interference across reasoning stages, and enable targeted optimization for heterogeneous tasks \cite{zhang2024chain,gao2025single,wang2025automisty,hong2023metagpt,huang2026tracecoder,wang2026mistypilot}. While prior MAS research has mainly focused on general planning, coding, and task-solving settings, the use of MAS for ASD intervention strategy detection in shared book reading interactions remains underexplored.

\paragraph*{Challenges in Multimodal Perception.}
Fine-grained analysis of parent-child shared book reading requires accurate speech recognition and reliable temporal grounding. However, child speech remains challenging for traditional ASR systems because of its substantial acoustic variability and pronunciation instability \cite{gerosa2007acoustic,horii2025children}. In parallel, recent multimodal large language models have shown promising video understanding ability, but they still struggle to provide precise temporal localization in long-form videos \cite{wu2025number,zhao2024needle}, especially when second-level timestamps are required for segment-level analysis. These limitations make robust multimodal perception a key bottleneck for automated intervention detection in naturalistic home environments.

\paragraph*{ASD Intervention Detection:}
In the domain of ASD language intervention assessment, the ASD-HI dataset \cite{li2025asd} provides the main benchmark for caregiver intervention strategy detection in home-based shared book reading videos.The strongest publicly available baseline uses Whisper for speech transcription and a single LLM agent guided by prompts designed by ASD intervention experts for intervention identification, followed by greedy sequential scanning for temporal segmentation. This benchmark establishes an important foundation for the task; however, its overall detection performance still leaves substantial room for improvement because it relies on error-prone transcripts and rigid heuristic boundary search.

\section{Problem Formulation}

We formulate the task as a strategy detection problem over parent–child shared book reading videos. Given an input video, the objective is to temporally localize intervention strategies and classify each detected segment into its corresponding strategy category.
Let $V \in \mathcal{V}$ denote a parent-child shared book reading video, and let $\mathcal{S}=\{\textit{Modeling},\textit{Mand\mbox{-}Model},\textit{Time Delay}\}$ be the intervention strategy space. The goal is to predict a set of temporally localized intervention strategy segments:
\begin{equation}
    \hat{\mathcal{Y}}=\{(\hat{t}_m^s,\hat{t}_m^e,\hat{s}_m)\}_{m=1}^{M}, \quad \hat{s}_m\in\mathcal{S},
\end{equation}
where $M$ is the total number of predicted interventions. For the $m$-th predicted segment, $\hat{t}_m^s$ and $\hat{t}_m^e$ denote its predicted start and end timestamps, respectively, and $\hat{s}_m$ represents its assigned strategy label.

InterventionLens first utilizes the OPA module to process the raw video $V$ into grounded interaction transcript features $\mathcal{X}$. Subsequently, the BMA module is applied to extract the book context representation $\mathcal{B}$ aligned with the current interaction:
\begin{equation}
    \mathcal{X}=\textit{OPA}(V), \qquad \mathcal{B}=\textit{BMA}(\mathcal{X}).
\end{equation}

Next, the system employs the ICSA module to generate a set of $K$  intervention candidates along with their routing cues based on the transcript features $\mathcal{X}$:
\begin{equation}
    \mathcal{C}=\{(c_k,q_k)\}_{k=1}^{K}=\textit{ICSA}(\mathcal{X}),
\end{equation}
where $c_k=(t_k^s, t_k^e)$ denotes the $k$-th intervention candidate segment with its initial coarse start and end times, and $q_k \in \mathcal{Q}=\{\textit{MODEL\mbox{-}like},\textit{MAND\mbox{-}like},\textit{TD\mbox{-}like}\}$ is the corresponding strategy cue used for expert routing.

Each candidate is then routed to the corresponding strategy-specific expert $\textit{Expert}_{q_k}$ for verification and boundary refinement. The expert first evaluates the candidate $c_k$ against its predefined strategy criteria in conjunction with the book context $\mathcal{B}$ to verify and assign the definitive strategy label $\hat{s}_k \in \mathcal{S}$. For accepted candidates, it subsequently adjusts the coarse boundaries $(t_k^s, t_k^e)$ into precise boundaries $(\hat{t}_k^s, \hat{t}_k^e)$:
\begin{equation}
    \hat{y}_k = \textit{Expert}_{q_k}(c_k,\mathcal{B}),
\end{equation}
where the output is explicitly defined as $\hat{y}_k \in \{(\hat{t}_k^s, \hat{t}_k^e, \hat{s}_k), \varnothing\}$. Here, $\varnothing$ denotes that the candidate is rejected (i.e., classified as background or invalid) during the strategy-specific verification.

The final prediction set is composed of all accepted and temporally refined candidates:
\begin{equation}
    \hat{\mathcal{Y}}=\{\hat{y}_k \mid \hat{y}_k \neq \varnothing\}_{k=1}^{K}.
\end{equation}

\section{Method}
\label{sec:method}
\begin{figure*}[t]
    \centering
    \includegraphics[width=\textwidth]{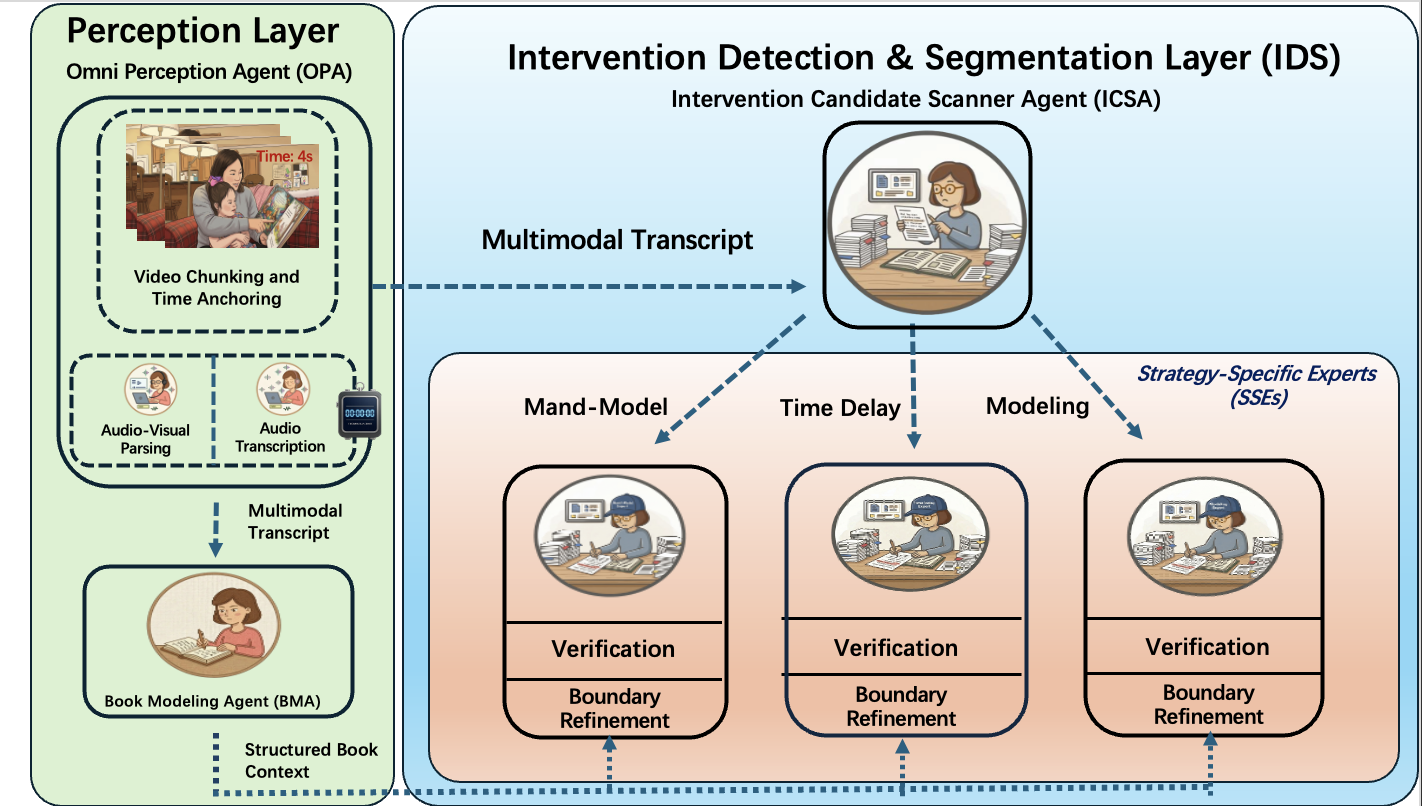}
    \caption{Overview of InterventionLens. The framework consists of a Perception Layer and an Intervention Detection \& Segmentation Layer (IDS). The Perception Layer first converts the raw shared book reading video into a time-aligned multimodal transcript and structured book context. The IDS Layer then proposes candidate intervention segments and routes them to strategy-specific experts, each of which performs strategy-specific verification and boundary refinement to produce the final temporally localized intervention segments.}
    \label{fig:main_method}
\end{figure*}

As shown in Fig.~\ref{fig:main_method}, InterventionLens is a two-layer multi-agent framework that decouples multimodal perception and contextual grounding from downstream intervention detection and boundary refinement. Specifically, the Perception Layer converts raw shared book reading videos into time-aligned multimodal interaction transcripts and infers structured book context from the ongoing session. Conditioned on these structured representations, the Intervention Detection and Segmentation (IDS) Layer follows a coarse-to-fine pipeline that first extracts and routes candidate intervention segments, and then performs strategy-specific verification and temporal boundary refinement.

\subsection{Perception Layer}

The Perception Layer consists of two components: the Omni-Perception Agent (OPA), which constructs time-aligned multimodal interaction transcripts, and the Book Modeling Agent (BMA), which reconstructs a structured book-context representation.

\paragraph*{Omni-Perception Agent (OPA):}

The Omni-Perception Agent (OPA) constructs time-aligned multimodal interaction transcripts from caregiver-child Shared Book Reading (SBR) videos for downstream intervention analysis. It captures verbal exchanges between caregivers and children together with observable actions and facial expressions from both participants.
We use \textit{gemini-3-flash-preview} as the backbone multimodal audio-video understanding model to parse dialog and visible interaction cues. For temporal grounding, we follow an OCR-assisted timestamping strategy in which a red timestamp is overlaid in the bottom-right corner of each frame and used as an explicit temporal reference to predict start and end times~\cite{wu2025number}. For long recordings, each reading session is processed as a sequence of 15-second clips.
In parallel, the audio stream is transcribed using \textit{Whisper}. We then align Gemini-generated caregiver utterances with Whisper transcriptions under semantic and temporal consistency constraints. When a Gemini-parsed caregiver utterance has a semantically matched Whisper counterpart with closely aligned timestamps, we replace the Gemini-predicted temporal boundaries with the corresponding Whisper timestamps. The resulting transcript serves as perceptual input to the downstream BMA and IDS Layer.

\paragraph*{Book Modeling Agent (BMA):}
Shared book reading forms a triadic interaction among the caregiver, the child, and the book. As a result, caregiver interventions are often grounded in the specific context of the book being read rather than in dialogue alone.
To incorporate this contextual information, we introduce the Book Modeling Agent (BMA), which infers a Structured Book-Context Representation from the parent--child interaction transcript. Instead of reconstructing the full book content, BMA extracts contextual cues such as recurring sentence patterns and potential target words that are relevant to the current reading interaction.
This inferred book context is then used as semantic grounding for downstream expert modules, enabling more accurate intervention detection by aligning caregiver utterances with the underlying reading context.

\subsection{Intervention Detection \& Segmentation Layer }
The Intervention Detection and Segmentation Layer detects and temporally segments caregiver intervention strategies by leveraging the enriched interaction transcripts and reconstructed book context produced by the Perception Layer. It is implemented through task-specific agents whose decision logic is derived from the ASD-HI coding manual and refined on the training split. Rather than training model parameters, we translate the operational definitions of Modeling, Mand-Model, and Time Delay into reusable decision rules.

\paragraph*{Intervention Candidate Scanner Agent (ICSA)}
ICSA is a module for candidate intervention extraction. Based on the caregiver-child interaction transcripts produced by the Perception Layer, it identifies interaction segments that may constitute intervention events. Specifically, ICSA extracts two clinically meaningful candidate patterns: Complete Intervention Loops, which follow a caregiver stimulus, child response, and caregiver reinforcement sequence, and Intervention Attempts, in which a caregiver provides a stimulus but the child does not respond. Based on these candidates, ICSA further assigns a coarse-grained strategy cue, such as MODEL-like, MAND-like, or TD-like, and uses it as a routing signal to dispatch each candidate to the corresponding strategy-specific expert for subsequent fine-grained verification and boundary refinement.
\paragraph*{Strategy-Specific Experts (SSEs)}
The Strategy-Specific Experts (SSEs) perform strategy-specific verification and temporal boundary refinement for the candidate intervention segments routed from the Intervention Candidate Scanner Agent (ICSA). To account for the substantial differences in linguistic patterns and interaction structures among Modeling, Mand-Model, and Time Delay, we design SSE as a parallel strategy-specific expert architecture. This architecture consists of three independent expert systems, each dedicated to one intervention strategy. All three expert systems follow the same two-stage cascade paradigm, where Stage 1 performs strategy-specific verification and Stage 2 refines temporal boundaries.

\noindent\hspace{1em}\textit{Strategy-Specific Verification:}
Strategy-Specific Verification is the core logical filtering stage for candidate intervention screening. Because the candidate segments proposed by ICSA still contain noisy or weak matches, this stage verifies each candidate using strategy-specific constraints induced from the ASD-HI dataset together with book-context cues from the Book Modeling Agent (BMA). The verification considers utterance form, interaction structure, local discourse context, and book-related patterns. Only candidates consistent with the target strategy are retained for subsequent boundary refinement.

\noindent\hspace{1em}\textit{Temporal Boundary Refinement:}
After strategy verification, the retained candidates proceed to the temporal boundary refinement stage, where their temporal extent is localized more precisely. Guided by local dialogue flow and expert annotation conventions, this stage refines the start and end boundaries of each intervention segment. Its goal is to narrow the relatively broad candidate region from the previous stage into a precise intervention interval by removing adjacent turns before and after the core intervention that do not belong to the intervention itself. As a result, the refined segments are more closely aligned with expert annotation standards and more accurately capture the actual temporal span of the intervention.

\subsection{Progressive Knowledge Refinement for SSEs}
We initialize the agent’s structured guidance using the ASD-HI annotation protocol, which specifies strategy definitions, temporal boundary judgment, and key interaction cues. We then iteratively refine the guidance on the training split with F1 score as the optimization objective. In each iteration, we run the full system, analyze recurrent failure cases, and update the guidance rules based on the most frequent error patterns. We stop the refinement process when the F1 score does not improve for three consecutive iterations.

\section{Experiments}
\label{sec:experiments}
This section presents the experimental evaluation of InterventionLens on the ASD-HI benchmark. To ensure direct comparability, we follow the same evaluation protocol as prior work. We first introduce the dataset, evaluation metrics, and implementation details. We then compare InterventionLens with the strongest reported baseline and conduct an ablation on the Book Modeling Agent (BMA). Finally, we provide additional analysis on the unguided pre-feedback subset, where caregivers had not yet received expert feedback, and discuss the main error patterns observed in the benchmark.

\subsection{Experimental Setup}
\paragraph*{Dataset}
We evaluate InterventionLens on the ASD-HI \cite{li2025asd} benchmark, a multimodal parent-child shared book reading dataset collected in home-based intervention settings for families of children with ASD. The dataset records natural caregiver child interactions in real home environments. It contains 48 complete shared-reading sessions from 3 families, together with expert-provided fine grained annotations of caregiver intervention segments. For each intervention segment, the dataset provides not only the corresponding intervention strategy label and its temporal boundaries, but also fine-grained annotations of the caregiver-child interaction process within the segment, including dialogue content, behavioral actions, and other interaction cues. In addition, the samples in the dataset can be further divided into two groups: those collected before caregivers received expert feedback, and those collected after caregivers received expert feedback.

Due to variations in home environments and recording conditions, camera placement is not consistent across families. Moreover, caregivers differ in prompting style, and children vary in engagement and responsiveness, resulting in substantial cross-family variability and distribution shift.The dataset provides 239 annotated segments for training and 120 annotated segments for evaluation. We note that 3 segments in the evaluation split contain incomplete annotations, specifically missing strategy labels or missing temporal boundaries. We therefore conduct our final evaluation on 117 valid test segments. All reported results follow the same evaluation split used by the original dataset authors.

\paragraph*{Evaluation Protocol}
We employ Precision, Recall, and F1-score as the primary evaluation metrics. A predicted intervention is counted as a true positive only if it simultaneously satisfies three core criteria: (1) Strategy Correctness, where the predicted strategy category must exactly match the expert-annotated ground truth; (2) Strategy Completeness, where the prediction must capture the complete caregiver child interaction process underlying the intervention, including all necessary interactional details, rather than only a fragmented or isolated portion of it; and (3) Temporal Accuracy, where the predicted time interval must align with the annotated reference boundaries within a reasonable tolerance window of 1.0 second. To assess the stability of model generation, we conduct three independent runs on the same evaluation split. For each run, we compute the results for the three strategy categories Mand-Model, Modeling, and Time Delay and report the macro-averaged performance. The final results are presented as the mean $\pm$ standard deviation across the three runs.

\paragraph*{Implementation Details}
In the perception layer, we use \textit{Whisper-1} for audio transcription and \textit{text-embedding-3-small} to support semantic alignment between Gemini parsed caregiver utterances and Whisper transcription segments. Specifically, alignment is performed by computing the cosine distance in the embedding space. We replace the Gemini-predicted temporal boundaries with the corresponding Whisper boundaries only when the cosine distance is below 0.1 and the boundary discrepancy is within 1.0 second; otherwise, the original Gemini timestamps are retained. For multimodal understanding and downstream reasoning, all agent components use \textit{gemini-3-flash-preview} as the backbone model, with the temperature set to 1.0.

During the Progressive Knowledge Refinement for SSEs stage, all rule induction and prompt refinement are performed on the subset collected after caregivers received expert feedback, consisting of 197 annotated segments. We use this subset because intervention behaviors at this stage are more standardized, with more stable strategy boundaries and interaction patterns, which makes it better suited for extracting reusable decision rules. In contrast, the interactions collected before expert feedback are more natural and variable, and are used in our subsequent analysis. We conduct nine rounds of refinement and stop the process when the F1 score fails to improve for three consecutive iterations, yielding a final training-set F1 score of 0.774.

\subsection{Main Results}

\begin{table}[t]
\centering
\caption{Overall results on ASD-HI}
\label{tab:main_overall}
\renewcommand{\arraystretch}{1.08}
\setlength{\tabcolsep}{4pt}
\small
\begin{tabular}{lccc}
\toprule
\textbf{Method} & \textbf{Prec.} & \textbf{Rec.} & \textbf{F1} \\
\midrule
Baseline & 50.21 & 73.68 & 59.72 \\
Ours & 82.36$\pm$3.44 & 76.75$\pm$0.49 & 79.44$\pm$1.84 \\
\midrule
Gain & +32.15 & +3.07 & +19.72 \\
\bottomrule
\end{tabular}
\end{table}

\begin{table}[t]
\centering
\caption{Strategy wise results on ASD-HI.}
\label{tab:main_strategy}
\renewcommand{\arraystretch}{1.08}
\setlength{\tabcolsep}{4pt}
\small
\begin{tabular}{lccc}
\toprule
\textbf{Strategy} & \textbf{Prec.} & \textbf{Rec.} & \textbf{F1} \\
\midrule
Mand-Model & 91.87$\pm$1.19 & 87.88$\pm$0.75 & 89.82$\pm$0.31 \\
Modeling   & 57.34$\pm$12.72 & 26.67$\pm$6.67 & 36.28$\pm$8.25 \\
Time Delay & 65.03$\pm$6.44 & 72.84$\pm$5.66 & 68.65$\pm$5.63 \\
\bottomrule
\end{tabular}

\vspace{1ex}
\parbox{0.95\linewidth}{%
\footnotesize
\textit{Note:} The evaluation contains 117 annotated intervention strategies, including 74 \textit{Mand-Model}, 27 \textit{Time Delay}, and 16 \textit{Modeling} instances.
}
\end{table}

We first compare InterventionLens with the baseline reported in the original ASD-HI paper under the same evaluation protocol. Since the original ASD-HI paper reports only overall performance without strategy-wise Precision, Recall, or F1 scores, we restrict our comparison to the overall metrics. As shown in Table~\ref{tab:main_overall}, InterventionLens substantially improves overall performance over the reported baseline~\cite{li2025asd}. Specifically, our method increases Precision from 50.21 to 82.36, Recall from 73.68 to 76.75, and F1 from 59.72 to 79.44. The most notable gain comes from the large improvement in \textbf{Precision} (\textbf{+32.15}), while \textbf{Recall} is also slightly improved (\textbf{+3.07}). Consequently, the overall \textbf{F1} score increases by (\textbf{+19.72}), indicating a substantially better balance between prediction accuracy and coverage. This suggests that InterventionLens does not achieve better performance simply by producing more aggressive predictions; instead, it substantially reduces false positives while maintaining strong coverage of true intervention segments.

Table~\ref{tab:main_strategy} further reports the per-strategy performance of InterventionLens. Our method achieves the strongest results on \textit{Mand-Model}, with an F1 of 89.82, suggesting that this strategy can be detected reliably under our framework. Performance on \textit{Time Delay} remains competitive, reaching an F1 of 68.65. In contrast, \textit{Modeling} is the most challenging category, with an F1 of 36.28. This gap suggests that different intervention strategies pose substantially different levels of difficulty, and that \textit{Modeling} remains the major bottleneck in the current benchmark.

In addition, as a supplementary analysis, we further examine the performance of InterventionLens on the unguided interaction subset collected before caregivers received expert feedback. On this subset, the system achieves an overall F1 of 85.09$\pm$2.24, suggesting that the rules induced from post-feedback data remain applicable to more natural family shared-reading interactions. However, this result should not be over interpreted, since the evaluation subset contains only 24 annotated segments and is highly imbalanced across strategy categories, with only 15, 6, 3 instances for Mand-Model, Time Delay, and Modeling, respectively. We therefore treat this result as a supplementary observation on system applicability, rather than as a definitive validation of cross-distribution generalization.

\subsection{Ablation Study}

\begin{table}[t]
\centering
\caption{Overall results under BMA ablation.}
\label{tab:ablation_overall}
\renewcommand{\arraystretch}{1.08}
\setlength{\tabcolsep}{4pt}
\small
\begin{tabular}{lccc}
\toprule
\textbf{Method} & \textbf{Prec.} & \textbf{Rec.} & \textbf{F1} \\
\midrule
InterventionLens & 82.36$\pm$3.44 & 76.75$\pm$0.49 & 79.44$\pm$1.84 \\
w/o BMA & 70.22$\pm$3.49 & 67.79$\pm$1.28 & 68.97$\pm$2.35 \\
\midrule
Drop & -12.14 & -8.96 & -10.47 \\
\bottomrule
\end{tabular}
\end{table}

\begin{table}[t]
\centering
\caption{Strategy-wise F1 under BMA ablation.}
\label{tab:ablation_strategy}
\renewcommand{\arraystretch}{1.08}
\setlength{\tabcolsep}{4pt}
\small
\begin{tabular}{lccc}
\toprule
\textbf{Strategy} & \textbf{InterventionLens} & \textbf{w/o BMA} & \textbf{$\Delta$F1} \\
\midrule
Mand-Model & 89.82$\pm$0.31 & 80.27$\pm$2.90 & -9.55 \\
Modeling   & 36.28$\pm$8.25 & 16.96$\pm$9.06 & -19.32 \\
Time Delay & 68.65$\pm$5.63 & 64.93$\pm$2.17 & -3.72 \\
\bottomrule
\end{tabular}

\end{table}

\paragraph{Ablation on the Book Modeling Agent (BMA).}
To evaluate the contribution of book-conditioned semantic grounding, we remove the Book Modeling Agent (BMA) while keeping all other components and experimental settings unchanged. As shown in Table~\ref{tab:ablation_overall}, removing BMA leads to a clear overall performance drop: Precision decreases from 82.36 to 70.22, Recall decreases from 76.75 to 67.79, and F1 decreases from 79.44 to 68.97. This result indicates that explicit book modeling is a key component of the full system rather than an optional auxiliary module.

Table~\ref{tab:ablation_strategy} further shows that the impact of removing BMA is not uniform across strategy categories. In particular, the F1 of Modeling drops from 36.28 to 16.96, while Mand-Model also decreases from 89.82 to 80.27. In contrast, Time Delay is relatively less affected, with F1 decreasing from 68.65 to 64.93. These results suggest that BMA contributes broadly to intervention strategy detection, and may be particularly important for strategies that require aligning caregiver utterances with book content and target-word semantics. We note that the Modeling category has relatively few instances in the evaluation split, and its category-level results should therefore be interpreted with caution; however, the clear degradation in both the overall results and Mand-Model still consistently indicates that explicit book modeling provides critical contextual grounding for the system. Overall, this ablation supports our central assumption that shared book reading should be modeled as a triadic interaction among the caregiver, the child, and the book.

\subsection{Error Analysis}
A notable observation is that the detection performance of \textit{Modeling} is substantially lower than that of the other two strategy categories. We believe that this result is influenced by at least two factors. First, the number of \textit{Modeling} instances in the evaluation split is very small, with only 16 samples, making the corresponding metrics more sensitive to fluctuations caused by a few individual cases. Second, we observe that there may be some inconsistency in labeling criteria between the training and evaluation splits. Specifically, when the same caregiver reads the same story book, \textit{Brown Bear, Brown Bear, What Do You See?}, utterances that are highly similar in both form and semantics are not always annotated consistently across splits. For example, in the training split, repeated target-word expressions such as ``black sheep \ldots black sheep'' are labeled as \textit{Modeling}, whereas in the evaluation split, similarly structured expressions from the same caregiver, such as ``red bird \ldots red bird'', are not labeled as \textit{Modeling}. This suggests that the \textit{Modeling} category is challenged not only by limited sample size, but also potentially by ambiguous labeling boundaries or inconsistent annotation criteria.

\section{Conclusion and Limitations}

In this paper, we presented InterventionLens, a hierarchical multi-agent framework for detecting and temporally segmenting caregiver intervention strategies in home-based shared reading videos. Unlike approaches that rely on task-specific parameter updating, we decompose the task into a sequence of interconnected subprocesses, including multimodal perception, book-context modeling, intervention candidate discovery, strategy-specific verification, and temporal boundary refinement. This task decomposition allows the system to organize the overall reasoning pipeline from perception to decision-making in a modular manner, while enabling fine-grained temporal localization of intervention segments on the current benchmark.

Experiments on the ASD-HI benchmark show that InterventionLens substantially outperforms the strongest reported baseline, achieving clear gains in overall Precision and F1. Further ablation results demonstrate that explicit book-conditioned semantic grounding is a key component of the full system rather than an optional auxiliary design. Modeling shared reading as a triadic interaction among the caregiver, the child, and the book provides important contextual support for intervention detection, and is particularly helpful for strategies that depend on aligning caregiver utterances with recurring story patterns and target-word semantics.

It should also be noted that InterventionLens was developed based on the ASD-HI training set and its corresponding task formulation. As a result, the method may still exhibit a certain degree of dependence on the benchmark's annotation conventions, strategy boundaries, and data distribution. Although the experimental results show that InterventionLens achieves substantial improvements under the current evaluation protocol, its transferability and robustness across datasets, family settings, and annotation schemes still require more systematic validation.

\section{Ethical Considerations}
\label{sec:ethics}
This study utilizes the ASD-HI benchmark dataset, which was collected through a prior Institutional Review Board (IRB)-approved study. The current research does not involve any new human-subject data collection; the released dataset is used strictly for academic research purposes in line with the approved data use guidelines.

For multimodal analysis, we accessed the Gemini model through Google Cloud Vertex AI. Customer data is not used to train Google's AI models as stated in their terms of use, and all API communications are protected via encryption in transit using TLS. Furthermore, we have strictly limited the use of data to research analysis, with no attempts to identify or re-identify participants, and no redistribution of raw videos or identifiable materials. 

{
    \small
    \bibliographystyle{ieeenat_fullname}
    \bibliography{main}

@book{american2013diagnostic,
  title={Diagnostic and statistical manual of mental disorders},
  author={American Psychiatric Association and others},
  year={2013},
  publisher={American psychiatric association}
}

@article{fuller2020effects,
  title={The effects of early intervention on social communication outcomes for children with autism spectrum disorder: A meta-analysis},
  author={Fuller, Elizabeth A and Kaiser, Ann P},
  journal={Journal of autism and developmental disorders},
  volume={50},
  number={5},
  pages={1683--1700},
  year={2020},
  publisher={Springer}
}

@article{estes2015long,
  title={Long-term outcomes of early intervention in 6-year-old children with autism spectrum disorder},
  author={Estes, Annette and Munson, Jeffrey and Rogers, Sally J and Greenson, Jessica and Winter, Jamie and Dawson, Geraldine},
  journal={Journal of the American Academy of Child \& Adolescent Psychiatry},
  volume={54},
  number={7},
  pages={580--587},
  year={2015},
  publisher={Elsevier}
}

@article{lin2025inequality,
  title={Inequality and heterogeneity in medical resources for children with autism spectrum disorders: a study in the ethnic minority region of southern China},
  author={Lin, Yingying and Chen, Guozhi and Lu, Huaxiang and Qin, Rongfei and Jiang, Jinsheng and Tan, Weiwei and Luo, Caibin and Chen, Ming and Huang, Qin and Huang, Liangliang and others},
  journal={BMC Public Health},
  volume={25},
  number={1},
  pages={1677},
  year={2025},
  publisher={Springer}
}

@article{lord2022lancet,
  title={The Lancet Commission on the future of care and clinical research in autism},
  author={Lord, Catherine and Charman, Tony and Havdahl, Alexandra and Carbone, Paul and Anagnostou, Evdokia and Boyd, Brian and Carr, Themba and De Vries, Petrus J and Dissanayake, Cheryl and Divan, Gauri and others},
  journal={The Lancet},
  volume={399},
  number={10321},
  pages={271--334},
  year={2022},
  publisher={Elsevier}
}

@article{world2022training,
  title={Training for caregivers of children with developmental disabilities, including autism},
  author={World Health Organization and others},
  journal={Retrieved January},
  volume={17},
  pages={2023},
  year={2022}
}

@article{wong2022adapting,
  title={Adapting and pretesting the World Health Organization’s Caregiver Skills Training Program for children with autism and developmental disorders or delays in Hong Kong},
  author={Wong, Paul Wai-Ching and Lam, Yan-Yin and Lau, Janet Siu-Ping and Fok, Hung-Kit and WHO CST Team Servili Chiara 4 Salomone Erica 5 6 Pacione Laura 4 5 Shire Stephanie 6 Brown Felicity 7 8},
  journal={Scientific Reports},
  volume={12},
  number={1},
  pages={16932},
  year={2022},
  publisher={Nature Publishing Group UK London}
}

@article{wetherby2018changing,
  title={Changing developmental trajectories of toddlers with autism spectrum disorder: Strategies for bridging research to community practice},
  author={Wetherby, Amy M and Woods, Juliann and Guthrie, Whitney and Delehanty, Abigail and Brown, Jennifer A and Morgan, Lindee and Holland, Renee D and Schatschneider, Christopher and Lord, Catherine},
  journal={Journal of Speech, Language, and Hearing Research},
  volume={61},
  number={11},
  pages={2615--2628},
  year={2018},
  publisher={American Speech-Language-Hearing Association}
}

@article{boyle2019effects,
  title={Effects of shared reading on the early language and literacy skills of children with autism spectrum disorders: A systematic review},
  author={Boyle, Susannah A and McNaughton, David and Chapin, Shelley E},
  journal={Focus on Autism and Other Developmental Disabilities},
  volume={34},
  number={4},
  pages={205--214},
  year={2019},
  publisher={Sage Publications Sage CA: Los Angeles, CA}
}

@article{westerveld2021investigating,
  title={Investigating the effectiveness of parent-implemented shared book reading intervention for preschoolers with ASD},
  author={Westerveld, Marleen F and Wicks, Rachelle and Paynter, Jessica},
  journal={Child Language Teaching and Therapy},
  volume={37},
  number={2},
  pages={149--162},
  year={2021},
  publisher={SAGE Publications Sage UK: London, England}
}

@inproceedings{li2025asd,
  title={ASD-HI: A Parent-Child Interaction Dataset for Automated Assessment of Home Intervention},
  author={Li, Zhaohui and Akemoglu, Yusuf and Lyu, Jincheng and Zheng, Qingxiao and Xiong, Jinjun},
  booktitle={International Conference on Artificial Intelligence in Education},
  pages={48--62},
  year={2025},
  organization={Springer}
}

@inproceedings{attia2024kid,
  title={Kid-whisper: Towards bridging the performance gap in automatic speech recognition for children vs. adults},
  author={Attia, Ahmed Adel and Liu, Jing and Ai, Wei and Demszky, Dorottya and Espy-Wilson, Carol},
  booktitle={Proceedings of the AAAI/ACM Conference on AI, Ethics, and Society},
  volume={7},
  pages={74--80},
  year={2024}
}

@article{jain2023adaptationwhispermodelschild,
  title={Adaptation of Whisper models to child speech recognition},
  author={Jain, Rishabh and Barcovschi, Andrei and Yiwere, Mariam and Corcoran, Peter and Cucu, Horia},
  journal={arXiv preprint arXiv:2307.13008},
  year={2023}
}

@article{akemoglu2022parent,
  title={A parent-implemented shared reading intervention via telepractice},
  author={Akemoglu, Yusuf and Hinton, Vanessa and Laroue, Dayna and Jefferson, Vanessa},
  journal={Journal of Early Intervention},
  volume={44},
  number={2},
  pages={190--210},
  year={2022},
  publisher={SAGE Publications Sage CA: Los Angeles, CA}
}

@inproceedings{wu2025number,
  title={Number it: Temporal grounding videos like flipping manga},
  author={Wu, Yongliang and Hu, Xinting and Sun, Yuyang and Zhou, Yizhou and Zhu, Wenbo and Rao, Fengyun and Schiele, Bernt and Yang, Xu},
  booktitle={Proceedings of the Computer Vision and Pattern Recognition Conference},
  pages={13754--13765},
  year={2025}
}

@article{zhao2024needle,
  title={Needle in a video haystack: A scalable synthetic evaluator for video mllms},
  author={Zhao, Zijia and Lu, Haoyu and Huo, Yuqi and Du, Yifan and Yue, Tongtian and Guo, Longteng and Wang, Bingning and Chen, Weipeng and Liu, Jing},
  journal={arXiv preprint arXiv:2406.09367},
  year={2024}
}

@article{guo2024large,
  title={Large language model based multi-agents: A survey of progress and challenges},
  author={Guo, Taicheng and Chen, Xiuying and Wang, Yaqi and Chang, Ruidi and Pei, Shichao and Chawla, Nitesh V and Wiest, Olaf and Zhang, Xiangliang},
  journal={arXiv preprint arXiv:2402.01680},
  year={2024}
}

@article{yang2024llm,
  title={Llm-based multi-agent systems: Techniques and business perspectives},
  author={Yang, Yingxuan and Peng, Qiuying and Wang, Jun and Wen, Ying and Zhang, Weinan},
  journal={arXiv preprint arXiv:2411.14033},
  year={2024}
}

@article{tran2025multi,
  title={Multi-agent collaboration mechanisms: A survey of llms, 2025},
  author={Tran, Khanh-Tung and Dao, Dung and Nguyen, Minh-Duong and Pham, Quoc-Viet and O’Sullivan, Barry and Nguyen, Hoang D},
  journal={URL https://arxiv. org/abs/2501.06322},
  year={2025}
}

@article{zhang2024chain,
  title={Chain of agents: Large language models collaborating on long-context tasks},
  author={Zhang, Yusen and Sun, Ruoxi and Chen, Yanfei and Pfister, Tomas and Zhang, Rui and Arik, Sercan},
  journal={Advances in Neural Information Processing Systems},
  volume={37},
  pages={132208--132237},
  year={2024}
}

@article{gao2025single,
  title={Single-agent or Multi-agent Systems? Why Not Both?},
  author={Gao, Mingyan and Li, Yanzi and Liu, Banruo and Yu, Yifan and Wang, Phillip and Lin, Ching-Yu and Lai, Fan},
  journal={arXiv preprint arXiv:2505.18286},
  year={2025}
}

@inproceedings{wang2025automisty,
  title={AutoMisty: a multi-agent LLM framework for automated code generation in the Misty social robot},
  author={Wang, Xiao and Dong, Lu and Rangasrinivasan, Sahana and Nwogu, Ifeoma and Setlur, Srirangaraj and Govindaraju, Venugopal},
  booktitle={2025 IEEE/RSJ International Conference on Intelligent Robots and Systems (IROS)},
  pages={9194--9201},
  year={2025},
  organization={IEEE}
}

@inproceedings{hong2023metagpt,
  title={MetaGPT: Meta programming for a multi-agent collaborative framework},
  author={Hong, Sirui and Zhuge, Mingchen and Chen, Jonathan and Zheng, Xiawu and Cheng, Yuheng and Wang, Jinlin and Zhang, Ceyao and Wang, Zili and Yau, Steven Ka Shing and Lin, Zijuan and others},
  booktitle={The twelfth international conference on learning representations},
  year={2023}
}

@article{huang2026tracecoder,
  title={TraceCoder: A Trace-Driven Multi-Agent Framework for Automated Debugging of LLM-Generated Code},
  author={Huang, Jiangping and Ye, Wenguang and Sun, Weisong and Zhang, Jian and Zhang, Mingyue and Liu, Yang},
  journal={arXiv preprint arXiv:2602.06875},
  year={2026}
}

@article{gerosa2007acoustic,
  title={Acoustic variability and automatic recognition of children’s speech},
  author={Gerosa, Matteo and Giuliani, Diego and Brugnara, Fabio},
  journal={Speech communication},
  volume={49},
  number={10-11},
  pages={847--860},
  year={2007},
  publisher={Elsevier}
}

@inproceedings{horii2025children,
  title={Why is children’s ASR so difficult? Analyzing children’s phonological error patterns using SSL-based phoneme recognizers},
  author={Horii, Koharu and Tawara, Naohiro and Ogawa, Atsunori and Araki, Shoko},
  booktitle={Proc. Interspeech},
  year={2025}
}

@article{akemouglu2022module,
  title={A module-based telepractice intervention for parents of children with developmental disabilities},
  author={Akemo{\u{g}}lu, Yusuf and Laroue, Dayna and Kudesey, Carolina and Stahlman, Mary},
  journal={Journal of Autism and Developmental Disorders},
  volume={52},
  number={12},
  pages={5177--5190},
  year={2022},
  publisher={Springer}
}

@article{akemoglu2021parent,
  title={A parent-implemented shared-reading intervention to promote communication skills of preschoolers with autism spectrum disorder},
  author={Akemoglu, Yusuf and Tomeny, Kimberly R},
  journal={Journal of Autism and Developmental Disorders},
  volume={51},
  number={8},
  pages={2974--2987},
  year={2021},
  publisher={Springer}
}

@article{meadan2013coaching,
  title={Coaching parents of young children with autism in rural areas using internet-based technologies: A pilot program},
  author={Meadan, Hedda and Meyer, Lori E and Snodgrass, Melinda R and Halle, James W},
  journal={Rural Special Education Quarterly},
  volume={32},
  number={3},
  pages={3--10},
  year={2013},
  publisher={SAGE Publications Sage CA: Los Angeles, CA}
}

@article{meadan2014parent,
  title={Parent-implemented social-pragmatic communication intervention: A pilot study},
  author={Meadan, Hedda and Angell, Maureen E and Stoner, Julia B and Daczewitz, Marcus E},
  journal={Focus on Autism and Other Developmental Disabilities},
  volume={29},
  number={2},
  pages={95--110},
  year={2014},
  publisher={Sage Publications Sage CA: Los Angeles, CA}
}

@article{meadan2016internet,
  title={Internet-based parent-implemented intervention for young children with autism: A pilot study},
  author={Meadan, Hedda and Snodgrass, Melinda R and Meyer, Lori E and Fisher, Kim W and Chung, Moon Y and Halle, James W},
  journal={Journal of Early Intervention},
  volume={38},
  number={1},
  pages={3--23},
  year={2016},
  publisher={Sage Publications Sage CA: Los Angeles, CA}
}

@article{wang2026mistypilot,
  title={MistyPilot: An Agentic Fast-Slow Thinking LLM Framework for Misty Social Robots},
  author={Wang, Xiao and Dong, Lu and Sun, Jingchen and Nwogu, Ifeoma and Setlur, Srirangaraj and Govindaraju, Venu},
  journal={arXiv preprint arXiv:2603.03640},
  year={2026}
}

@inproceedings{dong2024ig3d,
  title={Ig3d: Integrating 3d face representations in facial expression inference},
  author={Dong, Lu and Wang, Xiao and Setlur, Srirangaraj and Govindaraju, Venu and Nwogu, Ifeoma},
  booktitle={European Conference on Computer Vision},
  pages={404--421},
  year={2024},
  organization={Springer}
}

@inproceedings{dong2024word,
  title={Word-conditioned 3D American sign language motion generation},
  author={Dong, Lu and Wang, Xiao and Nwogu, Ifeoma},
  booktitle={Findings of the Association for Computational Linguistics: EMNLP 2024},
  pages={9993--9999},
  year={2024}
}

@inproceedings{dong2024signavatar,
  title={Signavatar: Sign language 3d motion reconstruction and generation},
  author={Dong, Lu and Chaudhary, Lipisha and Xu, Fei and Wang, Xiao and Lary, Mason and Nwogu, Ifeoma},
  booktitle={2024 IEEE 18th International Conference on Automatic Face and Gesture Recognition (FG)},
  pages={1--10},
  year={2024},
  organization={IEEE}
}
}

\end{document}